\documentclass{article}

\usepackage[numbers]{natbib}
\usepackage{graphicx}
\usepackage{placeins}
\usepackage{float}
\usepackage{wrapfig}
\usepackage{makecell}
\usepackage{amsmath}
\usepackage[preprint]{neurips_2026}

\usepackage[utf8]{inputenc} 
\usepackage[T1]{fontenc}    
\usepackage{hyperref}       
\usepackage{url}            
\usepackage{booktabs}       
\usepackage{amsfonts}       
\usepackage{nicefrac}       
\usepackage{microtype}      
\usepackage{xcolor}         

\title{Bridging Sequence and Graph Structure for Epigenetic Age Prediction}

\author{%
  Yao Li\\
  School of Computing and Information Systems\\
  The University of Melbourne\\
  \texttt{yao.li5@student.unimelb.edu.au} \\
   \And
   Xikun Zhang\thanks{corresponding author} \\
   School of Computing Technologies \\
   RMIT University \\
   \texttt{xikun.zhang@rmit.edu.au} \\
   \And
   Xiaotao Shen \\
   Lee Kong Chian School of Medicine \\
   Nanyang Technological University Singapore \\
   \texttt{xiaotao.shen@ntu.edu.sg} \\
   \And
   Sonika Tyagi \\
   School of Computing Technologies \\
   RMIT University \\
   \texttt{sonika.tyagi@rmit.edu.au} \\
   \And
   Xin Zheng \\
   School of Computing Technologies \\
   RMIT University \\
   \texttt{xin.zheng2@rmit.edu.au} \\
   \And
   Jiaxing Huang \\
   Department of Data Science and Artificial Intelligence \\
   Hong Kong Polytechnic University \\
   \texttt{jiaxing.huang0508@outlook.com} \\
   \And
   Feng Xia \\
   School of Computing Technologies \\
   RMIT University \\
   \texttt{f.xia@ieee.org} \\
}

\begin{document}

\maketitle

\begin{abstract}
Epigenetic clocks based on DNA methylation have emerged as
powerful tools for estimating biological age, with broad
applications in aging research, age-related disease studies,
and longevity science. Despite advances across machine learning approaches to epigenetic age prediction, spanning penalised linear regression, deep feedforward networks, residual architectures, and graph neural networks, no existing method jointly models co-methylation graph structure and site-specific DNA sequence context within a unified framework. We propose a unified sequence--graph integration framework for epigenetic age prediction that addresses this gap, integrating
eight-dimensional DNA sequence statistical features through
a lightweight gated modulation mechanism that adaptively
scales each site's methylation signal according to its
sequence-determined biological relevance prior to graph
convolution. Evaluated on 3,707 blood methylation samples
against a comprehensive set of baselines, our method
achieves a test MAE of 3.149 years, a 12.8\% improvement
over the strongest graph-based baseline. Biologically informed statistical features outperform CNN-based sequence encoding, demonstrating that handcrafted sequence features are more effective than end-to-end learned representations in this data regime. Post-hoc interpretability analysis identifies CpG density and local adenine frequency as features with age-dependent importance shifts, consistent with known mechanisms of age-related hypermethylation at CpG-dense promoter regions. Our code is at https://github.com/yaoli2022/graphage-seq.
\end{abstract}

\section{Introduction}

Aging is one of the most significant biological processes
impacting human health, yet its molecular mechanisms
remain incompletely understood. DNA methylation, predominantly the addition of methyl groups to cytosines at cytosine-guanine
dinucleotide (CpG) sites in mammals, is a major epigenetic
mechanism through which cells regulate gene expression
and maintain normal biological function~\cite{jaenischbird2003}. By controlling
which genes are transcriptionally active, methylation
patterns underpin fundamental cellular processes including
development, differentiation, and homeostasis. As
individuals age, these patterns undergo systematic and
reproducible changes, progressively disrupting the
regulatory programmes that sustain cellular function and
contributing to the emergence of age-related diseases
such as neurodegeneration, cardiovascular disease, and
cancer~\cite{horvath2013, hannum2013, lopezotin2013, lopezotin2023}. The close coupling between methylation
state and cellular functional status makes genome-wide
methylation profiles a powerful molecular readout of
biological age, and has motivated the development of
epigenetic clocks as tools for precise biological age
estimation, with broad applications in early disease
diagnosis, therapeutic monitoring, and longevity
research~\cite{horvathraj2018}.

Existing epigenetic clocks fall into two broad categories based on how they represent CpG sites. The first treats CpG sites as independent features, spanning penalised linear regression on selected sites~\cite{horvath2013, hannum2013} to deep learning architectures including feedforward networks and residual convolutional models that improve prediction accuracy by capturing non-linear methylation dynamics~\cite{galkin2021, delimacamillo2022}. Despite their improved accuracy, these methods discard the relational structure among CpG sites. The second category addresses this limitation by representing methylation data as a graph, where CpG sites are nodes and their relationships such as co-methylation, chromosomal proximity, and shared gene membership are edges, enabling graph neural networks to capture structural dependencies that flat feature vectors cannot~\cite{ahmed2025, ektefaie2023}. Despite this progress, no existing method jointly models co-methylation graph structure and site-specific DNA sequence context, leaving a fundamental gap that our work addresses.

Existing epigenetic clock models
share two fundamental limitations that have not been
addressed simultaneously. First, most methods treat CpG
sites as independent features, discarding the rich
relational structure among them. Co-methylation, the
tendency of spatially proximate or functionally related
CpG sites to exhibit correlated methylation states, encodes
biologically significant information about chromatin
organisation and transcriptional regulation that flat
feature vectors cannot capture~\cite{wu2019}. Second, no existing
method incorporates the local DNA sequence context of each
CpG site into its representation. The sequence environment
surrounding a CpG site is a primary determinant of its
methylation susceptibility: sequence-level properties such
as GC content, CpG density, and local nucleotide
composition govern DNA methyltransferase enzyme binding preferences~\cite{santoni2021, affinito2020}
and directly shape age-related methylation dynamics. These limitations are fundamentally coupled: 
without sequence context, the model cannot identify which 
relational structure is biologically meaningful, and without 
graph structure, sequence-informed signals cannot propagate 
across functionally related sites~\cite{digiovanni2023}.

We assert that a principled solution requires a unified
framework that jointly integrates co-methylation graph
structure and site-specific DNA sequence context. Graph
neural networks naturally capture relational structure by
enabling each CpG site to aggregate information from its
neighbours, while sequence features provide the site-specific biological
context that determines each site's regulatory significance.
The key challenge lies in effectively integrating these two sources of information: methylation signals vary across individuals and drive supervised learning, whereas sequence features are site-specific biological priors that are invariant across individuals and cannot be optimised through standard supervision, requiring a dedicated mechanism to bridge this information.

This work proposes a unified sequence--graph integration 
framework for epigenetic age prediction. CpG sites are 
represented as nodes in a graph with edges encoding 
co-methylation relationships, chromosomal proximity, and 
shared gene membership. For each node, an eight-dimensional 
statistical feature vector encoding GC content, CpG density, 
and local nucleotide composition is integrated through a 
lightweight gated modulation mechanism that adaptively scales 
each site's methylation signal according to its 
sequence-determined biological relevance before graph 
convolution. We further compare two sequence encoding 
strategies, demonstrating that biologically informed 
statistical features outperform CNN-based one-hot encoding 
in this data regime. Our main contributions are as follows:

\begin{itemize}
    \item \textbf{Joint sequence and graph integration.}
    We propose the first unified framework that simultaneously 
    models co-methylation graph structure and site-specific DNA 
    sequence context for epigenetic age prediction, achieving a 
    test MAE of 3.149 years and surpassing all evaluated baselines 
    including classical epigenetic clocks, deep learning predictors, 
    and graph-based methods.
    \item \textbf{Gated sequence modulation.} We introduce
    a lightweight gated mechanism that adaptively scales each 
    site's methylation signal according to its sequence-determined 
    biological relevance, yielding an 18.1\% MAE improvement over 
    the sequence-agnostic baseline and demonstrating that sequence 
    context provides complementary information beyond graph 
    structure alone.
    \item \textbf{Sequence encoding strategy comparison.}
    We demonstrate that biologically informed statistical features 
    outperform CNN-based one-hot encoding in this data regime, 
    showing that handcrafted features encoding established 
    determinants of methylation susceptibility are more effective 
    than end-to-end sequence learning when training data is limited.
    \item \textbf{Biological interpretability.} Post-hoc analysis 
    identifies CpG density and local adenine frequency as exhibiting 
    age-dependent importance shifts consistent with known mechanisms 
    of age-related hypermethylation, providing candidate biomarkers 
    for aging intervention research.
\end{itemize}

\section{Related Work}
\paragraph{Epigenetic clocks.}
The field of epigenetic age prediction was established by
landmark regression-based models that identified CpG sites
whose methylation correlates linearly with chronological
age~\cite{horvath2013, hannum2013}. Subsequent research 
demonstrated that deep learning architectures could more 
effectively capture non-linear methylation dynamics: 
DeepMAge~\cite{galkin2021} utilized a four-layer feedforward 
network trained on blood methylation data, whereas 
AltumAge~\cite{delimacamillo2022} extended this framework to 
a pan-tissue clock employing multilayer perceptrons and 
interpreting predictions through SHAP values. Shi et
al.~\cite{shi2024} proposed ResnetAge, a residual convolutional
network that treats the ordered sequence of CpG beta values
as a one-dimensional signal, achieving competitive
performance across multiple tissue types. More recently,
Prosz et al.~\cite{prosz2024} introduced XAI-AGE, a 
biologically informed, explainable deep neural network that 
incorporates pathway-level knowledge into its architecture, 
thereby illustrating that embedding biological structure 
within model design enhances both predictive accuracy and 
interpretability. Beyond chronological age prediction, 
second-generation clocks such as PhenoAge~\cite{levine2018} 
have demonstrated that incorporating clinical biomarkers 
into methylation-based models yields stronger predictions 
of mortality and healthspan. Despite these advancements, a 
common limitation persists across all these methodologies: 
CpG sites are treated as independent features, with neither
the relational structure among them nor the site-specific
DNA sequence context that governs their methylation
behaviour~\cite{schubeler2015} taken into account.

\paragraph{Graph-based methylation modelling.}
To address the independence assumption, recent research has
explored graph-based representations of methylation data.
Wu et al.~\cite{wu2019} developed co-methylation networks from RNA
methylation data and demonstrated that modeling pairwise
relationships among sites reveals functional associations
that are not apparent in flat feature vectors. Building
upon the graph paradigm, Ahmed et al.~\cite{ahmed2025} introduced the
first GNN-based epigenetic clock, named GraphAge, which
represents CpG sites as nodes within a graph connected by
co-methylation values, chromosomal proximity, and shared
gene membership, and employs a PNA convolutional
layer~\cite{corso2020} for age regression. GraphAge demonstrated that
integrating structural information among CpG sites improves
both predictive accuracy and biological interpretability.
Nevertheless, across all these graph-based approaches,
node representations depend solely on methylation beta
values and fixed positional attributes, leaving the local
DNA sequence context of each CpG site completely
unexploited~\cite{digiovanni2023}.

\paragraph{Sequence context in methylation susceptibility.}
A separate line of research has established that the local
DNA sequence environment is a primary determinant of CpG
methylation susceptibility. Angermueller et al.~\cite{angermueller2017}
proposed DeepCpG, a convolutional neural network-based
model designed to predict single-cell methylation states
from local sequence windows, demonstrating that sequence
motifs carry substantial predictive power for methylation
variability~\cite{nguyen2023, zhou2024, schiff2024, marin2024}. Santoni~\cite{santoni2021} further showed that flanking
sequence composition directly governs CpG methylation
patterns, while Affinito et al.~\cite{affinito2020} demonstrated that
nucleotide distance modulates co-methylation strength
between proximal sites. Collectively, these findings
establish that sequence context encodes complementary
regulatory information that methylation values alone
cannot capture~\cite{hu2024}.

Despite this evidence, no prior work has jointly integrated
site-specific DNA sequence context and co-methylation graph
structure within a unified epigenetic age prediction
framework. Graph-based models exploit relational structure
but ignore sequence context; sequence-aware models such as
DeepCpG address methylation susceptibility but operate in
the single-cell imputation setting and do not model
population-level graph structure. Our work bridges this gap
by proposing a unified sequence--graph integration
framework with a lightweight gated modulation mechanism,
systematically comparing handcrafted statistical features
against CNN-based one-hot encoding in this data-limited
setting, and evaluating against a comprehensive set of
baselines spanning classical epigenetic clocks, deep
learning age predictors, and graph-based methods.

\section{Methods}

We propose a unified sequence--graph integration framework
for epigenetic age prediction that jointly models 
co-methylation graph structure and site-specific DNA 
sequence context via a lightweight gated modulation 
mechanism, followed by a PNA convolutional 
layer~\cite{corso2020} and an MLP for age regression.

\subsection{DNA Sequence Feature Extraction}

For each CpG site, an eight-dimensional statistical feature
 vector is extracted from its surrounding genomic sequence.
In contrast to methylation beta values, which exhibit
inter-individual variability, these features represent
intrinsic physicochemical and compositional properties of
each CpG locus and are invariant across samples. Each
dimension is designed to capture a distinct and biologically
interpretable characteristic of the local sequence
environment, as detailed in Table~\ref{tab:seq_features}.

\begin{table}[H]
\caption{Eight-dimensional DNA sequence statistical features
         extracted for each CpG site.}
\label{tab:seq_features}
\centering
\small
\begin{tabular}{lll}
\toprule
Dimension & Feature & Description \\
\midrule
1 & GC content & Proportion of G and C nucleotides in the
                 full sequence \\
2 & CpG density & Frequency of CG dinucleotides relative to
                  sequence length \\
3 & Upstream GC content & GC proportion in the 60\,bp upstream
                          of the CpG site \\
4 & Downstream GC content & GC proportion in the downstream
                            region \\
5 & Local adenine frequency & Proportion of A in the 10\,bp
                              window surrounding the CpG site \\
6 & Local thymine frequency & Proportion of T in the 10\,bp
                              window surrounding the CpG site \\
7 & Local cytosine frequency & Proportion of C in the 10\,bp
                               window surrounding the CpG site \\
8 & Local guanine frequency & Proportion of G in the 10\,bp
                              window surrounding the CpG site \\
\bottomrule
\end{tabular}
\end{table}


\subsection{Gated Modulation Architecture}

\begin{figure}[t]
\centering
\makebox[\textwidth][l]{%
    \hspace*{0cm}
    \includegraphics[width=\linewidth]{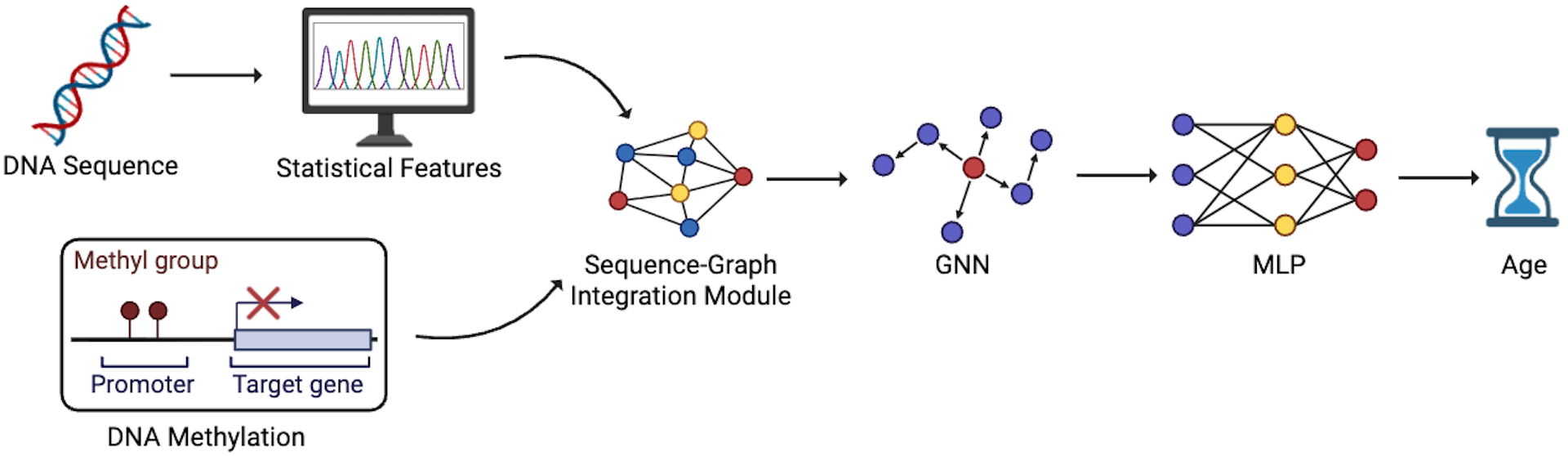}%
}
\caption{Overview of the proposed sequence--graph integration framework. Eight-dimensional DNA sequence features and methylation signals are fused via gated modulation, where methylation denotes methyl group addition at CpG sites in promoter regions that silences target genes. The fused representation is passed through a GNN and MLP for age regression.}
\label{fig:architecture}
\end{figure}

As illustrated in Figure~\ref{fig:architecture}, the model 
combines two parallel input pathways, one extracting 
eight-dimensional DNA sequence statistical features and 
the other carrying methylation signals with positional 
attributes, that converge in the Sequence-Graph Integration 
Module before being passed through a GNN and MLP for age 
regression.

The model operates on a graph of $N = 20{,}318$ CpG sites. 
Each node $i$ carries a ten-dimensional node feature vector 
comprising one methylation beta value and nine fixed 
positional attributes, and an eight-dimensional sequence 
statistical feature vector extracted from the 122\,bp\footnote{Base pair (bp) is the fundamental unit 
of DNA length.} surrounding genomic sequence. Each edge carries a 
three-dimensional attribute vector encoding co-methylation 
correlation, same-chromosome, and same-gene indicators. 
The sequence features are processed by two lightweight 
networks: an importance gate producing a scalar weight 
$g_i \in (0,1)$, and a projection network producing a 
two-dimensional representation $\mathbf{p}_i$. The modulated 
methylation value $\tilde{m}_i = m_i \cdot g_i$ is 
concatenated with the positional attributes and 
$\mathbf{p}_i$ to form a 12-dimensional fused node 
representation, passed through a PNA convolutional layer 
and an MLP for age prediction.

\paragraph{Importance Gating.}
Given the sequence feature vector $\mathbf{s}_i \in
\mathbb{R}^8$ for CpG site $i$, the importance gate
computes a scalar modulation weight through a two-layer
feedforward network with a sigmoid activation:
\begin{equation}
    g_i = \sigma\!\left(W_2 \cdot \text{ReLU}\!\left(
    W_1 \mathbf{s}_i + \mathbf{b}_1\right) + b_2\right)
\end{equation}
where $W_1 \in \mathbb{R}^{16 \times 8}$,
$\mathbf{b}_1 \in \mathbb{R}^{16}$,
$W_2 \in \mathbb{R}^{1 \times 16}$, $b_2 \in \mathbb{R}$
are learnable parameters, and $\sigma(\cdot)$ denotes the
sigmoid function. The original methylation beta value
$m_i \in \mathbb{R}$ is then modulated as:
\begin{equation}
    \tilde{m}_i = m_i \cdot g_i
\end{equation}
This formulation enables the model to differentially weight
methylation signals according to their sequence-determined
biological relevance, attenuating contributions from sites
whose local sequence context confers limited age-predictive
information while amplifying those associated with sequence
environments characteristic of strong age-related
methylation dynamics.

\paragraph{Sequence Projection.}
A secondary projection network maps the sequence feature
vector to a two-dimensional representation that encodes
complementary sequence-level information alongside the
scalar gate, providing an explicit sequence-derived
component within the fused node representation:
\begin{equation}
    \mathbf{p}_i = W_4 \cdot \text{ReLU}\!\left(
    W_3 \mathbf{s}_i + \mathbf{b}_3\right) + \mathbf{b}_4
\end{equation}
where $W_3 \in \mathbb{R}^{8 \times 8}$,
$\mathbf{b}_3 \in \mathbb{R}^{8}$,
$W_4 \in \mathbb{R}^{2 \times 8}$,
$\mathbf{b}_4 \in \mathbb{R}^2$ are learnable parameters.

\paragraph{Node Representation Fusion.}
The final node representation is constructed by
concatenating three components:
\begin{equation}
    \mathbf{x}_i = \left[\tilde{m}_i \;\|\; \mathbf{f}_i
    \;\|\; \mathbf{p}_i\right] \in \mathbb{R}^{12}
\end{equation}
where $\mathbf{f}_i \in \mathbb{R}^9$ denotes the
positional feature vector. This fusion explicitly combines
sequence-derived information ($\tilde{m}_i$ and
$\mathbf{p}_i$) with graph-level positional attributes
($\mathbf{f}_i$) into a unified node representation before
graph convolution, ensuring that both sequence context and
structural information jointly inform the message passing
process. The PNA aggregation for node $i$ is defined as:
\begin{equation}
    \mathbf{h}_i = \bigoplus_{(j,i) \in \mathcal{E}}
    \text{PNA}\!\left(\mathbf{x}_i, \mathbf{x}_j,
    \mathbf{e}_{ij}\right)
\end{equation}
where $\mathcal{E}$ denotes the edge set,
$\mathbf{e}_{ij} \in \mathbb{R}^3$ is the edge feature
vector between sites $i$ and $j$, and $\bigoplus$ denotes
the combination of multiple aggregators (mean, max, min,
and standard deviation), each scaled by identity,
amplification, and attenuation scalers as defined in the
original PNA formulation~\cite{corso2020}.

\paragraph{Age Regression.}
The output of the PNA layer $h_i \in \mathbb{R}$ is passed
through a ReLU activation, yielding a scalar embedding per
CpG site. The full embedding vector across all
$N = 20{,}318$ CpG sites is passed through a multilayer
perceptron (MLP) to produce the predicted age $\hat{y}_k$.
The model is trained by minimising the mean squared error
between predicted and true chronological age:
\begin{equation}
    \mathcal{L} = \frac{1}{N_s} \sum_{k=1}^{N_s}
    \left(\hat{y}_k - y_k\right)^2
\end{equation}
where $N_s$ is the number of training samples,
$\hat{y}_k$ is the predicted age, and $y_k$ is the true
chronological age of sample $k$. The MLP architecture and
training configuration are detailed in
Section~\ref{sec:setup}.

\section{Experiment}
\label{sec:setup}

\paragraph{Dataset.}
All experiments are conducted on healthy blood methylation
samples compiled from 37 publicly accessible datasets
obtained from the NCBI Gene Expression Omnibus
(GEO)~\cite{ncbi2024} and EBI ArrayExpress~\cite{ebi2024}.
All samples are measured using the Illumina
HumanMethylation27 platform
(GPL8490\footnote{\url{https://www.ncbi.nlm.nih.gov/geo/query/acc.cgi?acc=GPL8490}}),
which covers 20,318 CpG sites common across platforms. The
combined dataset comprises 3,707 healthy samples spanning
an age range of 0 to 93 years, with a gender distribution
of 37.4\% male and 62.6\% female. The age distribution is
depicted in Figure~\ref{fig:age_dist} (Appendix~\ref{app:dataset}).
From each individual dataset, 20\% of samples are reserved
as a held-out test set, resulting in 756 test samples. The
remaining samples are partitioned using 5-fold
cross-validation; training and evaluation are performed on
fold~2, yielding 2,360 training samples and 591 validation
samples. Full details of the CpG site annotation preprocessing, node 
feature construction, and graph construction procedure are 
provided in Appendix~\ref{app:dataset}.

\paragraph{Baselines.}
We evaluate two categories of baselines on the same
held-out test set under identical data conditions.

\paragraph{Epigenetic clock baselines.}
We include five established epigenetic clocks spanning
classical and deep learning approaches. Horvath~\cite{horvath2013} is a
penalised linear regression model applied using its
published coefficients. AltumAge~\cite{delimacamillo2022} is a pan-tissue
MLP-based clock applied using its published pre-trained
weights. DeepMAge~\cite{galkin2021} and ResnetAge~\cite{shi2024} are deep learning
clocks whose weights are not publicly available; we
re-implement their published architectures and retrain
both under identical experimental conditions to ensure a
fair comparison. GraphAge~\cite{ahmed2025} reports a PNA-GNN result on
the same platform and dataset split, included as an
external reference point.

\paragraph{Controlled ablation variants.} We compare three 
configurations differing only in sequence integration: 
PNA-GNN~\cite{ahmed2025, corso2020}, the sequence-agnostic 
baseline; PNA-GNN + CNN Sequence Encoding, using a 
three-layer 1D CNN to extract 32-dimensional sequence 
embeddings; and PNA-GNN + Statistical Features (Ours), 
using the eight-dimensional statistical features described 
above.

\paragraph{Evaluation Metrics.}
We report Mean Absolute Error (MAE) as the primary
evaluation metric~\cite{horvath2013, galkin2021, delimacamillo2022}. MAE provides a direct and
interpretable measure of prediction accuracy in units of
years and is less sensitive to outliers than Mean Squared
Error (MSE). We additionally report MSE and the coefficient
of determination ($R^2$) to characterise the overall
alignment between predicted epigenetic age and true
chronological age.

\paragraph{Implementation Details.}
The MLP consists of eight fully connected layers with
output dimensions 1024, 656, 256, 124, 64, 32, 8, and 1,
interleaved with ReLU and SELU~\cite{klambauer2017} activations and dropout
regularisation~\cite{srivastava2014} ($p = 0.2$) at selected layers. The model
is optimised using Adam~\cite{kingma2015} with learning rate $5 \times
10^{-4}$ and weight decay $5 \times 10^{-3}$, with a
ReduceLROnPlateau scheduler (patience = 4, factor = 0.4,
minimum learning rate $10^{-8}$) over 140 epochs. All
experiments are conducted on an NVIDIA H100 GPU. Full hardware and training details are provided in
Appendix~\ref{app:hardware}.

\FloatBarrier

\subsection{Comparison Against Epigenetic Clock Baselines}
\label{sec:results}

\begin{figure}[t]
\centering
\makebox[\textwidth][l]{%
    \hspace*{0cm}
    \includegraphics[width=\linewidth]{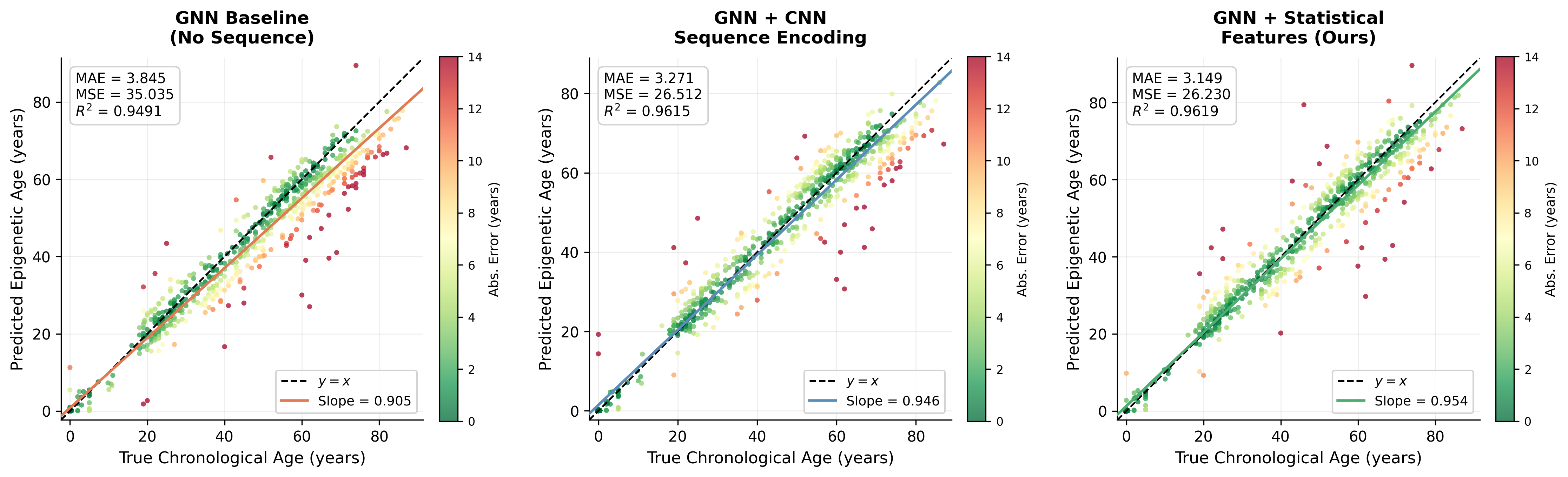}%
}
\caption{Predicted vs.\ true age for the three variants ($n = 756$).}
\label{fig:scatter}
\end{figure}

Table~\ref{tab:main_results} presents the test set
performance of all model configurations.
Figure~\ref{fig:scatter} visualises the alignment between
predicted and true age across the three controlled
sequence--graph integration variants.

\begin{table}[t]
\caption{Test set performance ($n = 756$).}
\label{tab:main_results}
\centering
\begin{tabular}{lccc}
\toprule
Model & MAE$\downarrow$ & MSE$\downarrow$ & $R^2$$\uparrow$ \\
\midrule
Horvath~\cite{horvath2013}                             & 5.823 & 78.104 & 0.8865 \\
AltumAge~\cite{delimacamillo2022}                            & 3.710 & 47.675 & 0.9307 \\
DeepMAge~\cite{galkin2021}                            & 3.222 & 29.574 & 0.9570 \\
ResnetAge~\cite{shi2024}                           & 5.552 & 58.481 & 0.9150 \\
GraphAge~\cite{ahmed2025}                           & 3.611 & 29.966 & 0.9564 \\
PNA-GNN~\cite{ahmed2025, corso2020}                         & 3.845 & 35.035 & 0.9491 \\
\midrule
PNA-GNN + CNN Sequence Encoding                & 3.271 & 26.512 & 0.9615 \\
\textbf{PNA-GNN + Statistical Features (Ours)} & \textbf{3.149} & \textbf{26.230} & \textbf{0.9619} \\
\bottomrule
\end{tabular}
\end{table}

\paragraph{Comparison with epigenetic clock baselines.}
Among the established epigenetic clocks evaluated on our
test set, the classical linear regression model of
Horvath~\cite{horvath2013} achieves MAE = 5.823, reflecting the
well-known limitation of penalised linear models in
capturing non-linear methylation--age relationships across
diverse datasets. Deep learning clocks reduce this error
substantially: AltumAge~\cite{delimacamillo2022} achieves MAE = 3.710, and
DeepMAge~\cite{galkin2021} achieves MAE = 3.222 when retrained on our
dataset. ResnetAge~\cite{shi2024}, a residual convolutional
architecture applied directly to the methylation beta
value sequence, achieves MAE = 5.552. This indicates that
one-dimensional convolutional processing of CpG site
ordering is less effective than MLP-based approaches at
this dataset scale, where the ordering of CpG sites does
not encode biologically meaningful spatial structure.
GraphAge~\cite{ahmed2025} achieves MAE = 3.611, 
demonstrating that graph-based modelling of co-methylation 
structure captures relational dependencies that are not 
exploited by sequence-based deep learning approaches such 
as ResnetAge~\cite{shi2024}. Our proposed framework surpasses all established
baselines with MAE = 3.149.

\paragraph{Contribution of sequence--graph integration.}
Under identical experimental conditions, jointly integrating DNA sequence statistical features with co-methylation graph structure reduces the test MAE from 3.845 to 3.149, an 18.1\% improvement over the sequence-agnostic baseline. This gain reflects the complementary nature of the two information sources: graph structure captures relational dependencies among CpG sites, enabling each site to contextualise its methylation signal relative to functionally related neighbours, while sequence features provide site-specific biological priors that determine which methylation signals are age-relevant. Without sequence context, the model relies entirely on graph structure to discriminate age-related methylation changes from individual variation. In addition, without graph structure, sequence features cannot propagate regulatory context across functionally related sites. The gated modulation mechanism bridges the two by learning to weight each site's methylation signal according to its sequence-determined biological relevance, enabling the integrated framework to selectively amplify signals from sites whose methylation dynamics are systematically coupled to the aging process.

\paragraph{Sequence encoding strategy.}
Comparing the two integration strategies, statistical
feature extraction (MAE = 3.149) outperforms CNN-based
one-hot encoding (MAE = 3.271). This result contradicts
the common expectation that end-to-end learned
representations should be more expressive than
hand-crafted features, and can be explained by two key
factors. First, the CNN encoder must learn to compress a
$122 \times 4$ one-hot representation into a meaningful
32-dimensional embedding from only 2,360 training samples.
Unlike methylation values, which vary across individuals
and provide a dense supervision signal, the sequence for
each CpG site is fixed and identical across all samples.
The CNN therefore receives no per-sample feedback about
which sequence patterns predict age, and can only learn
from the aggregate statistical relationship between
sequence motifs and methylation-based age prediction, a
considerably weaker learning signal at this dataset scale.
Second, the eight-dimensional statistical features directly
encode biologically established determinants of methylation
susceptibility, including GC content, CpG density, and
local nucleotide composition, reflecting known DNA methyltransferase enzyme
binding preferences~\cite{santoni2021, affinito2020, gao2020} and CpG island regulatory
context~\cite{deaton2011}. These features condense well-established
biochemical knowledge into a compact representation,
eliminating the need for the model to rediscover these
relationships from sequence data alone.

\FloatBarrier

\subsection{Age Group and Temporal Analysis}
\label{sec:analysis}

\paragraph{Age group performance.}

\begin{figure}[h]
\centering
\includegraphics[width=\linewidth]{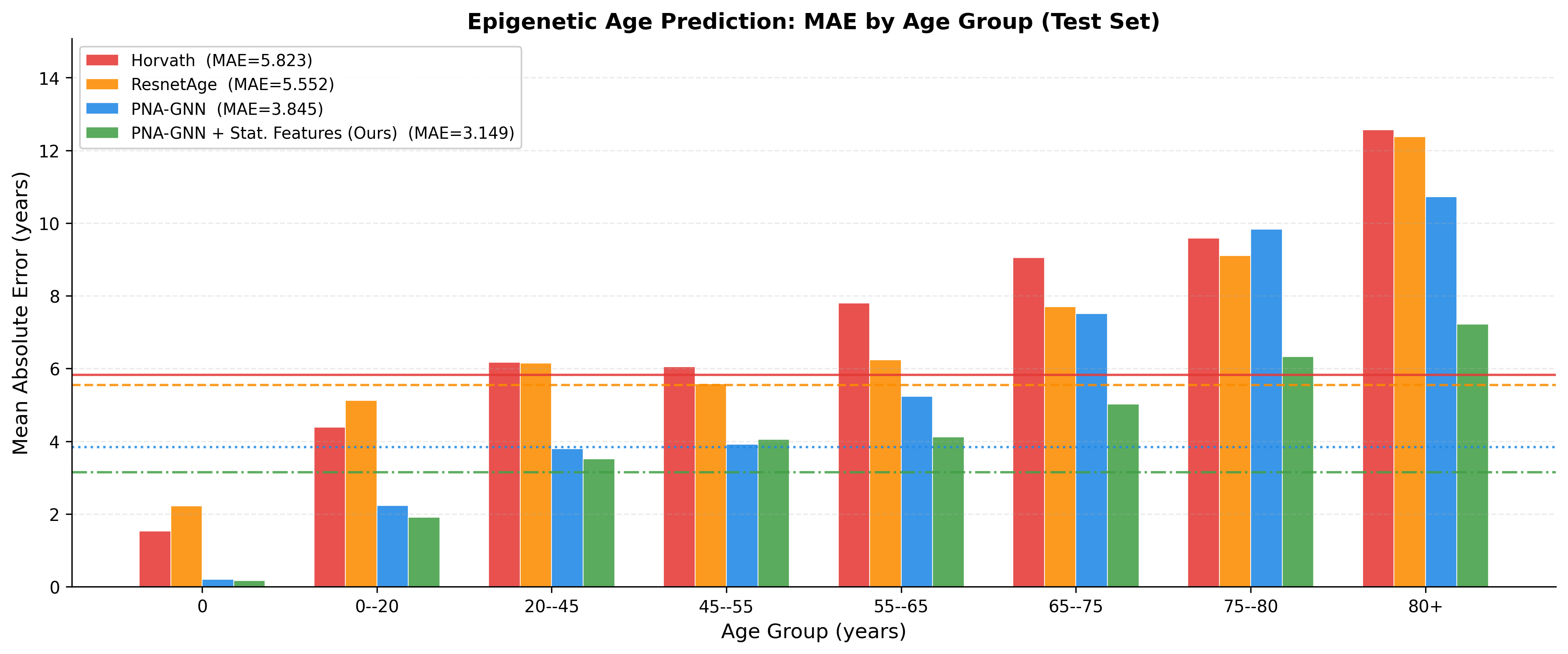}
\caption{Test MAE by age group ($n=756$).}
\label{fig:mae_age}
\end{figure}

Figure~\ref{fig:mae_age} presents the per-age-group MAE across four evaluated models. Our method achieves the lowest MAE in every age group, with prediction accuracy highest for neonatal subjects (MAE = 0.173 for age group 0) and degrading progressively with age, reaching MAE = 7.225 for subjects aged 80 and above. Horvath and ResnetAge show substantially higher errors across all age groups, while PNA-GNN remains competitive but is consistently outperformed by our method. This monotonic increase reflects a well-established biological phenomenon: in young subjects, epigenetic methylation patterns are tightly regulated and closely coupled to developmental programmes, resulting in a highly reproducible relationship between methylation and chronological age. As individuals age, stochastic epigenetic drift accumulates, causing methylation patterns to diverge progressively from the population mean~\cite{issa2014}. This drift is driven by the progressive loss of youthful epigenetic information as described by the Information Theory of Aging~\cite{lu2023}, imperfect maintenance of methylation marks during DNA replication, and the cumulative effect of environmental exposures, all of which introduce noise that cannot be captured by any methylation-based clock~\cite{jones2015}. The consequence is an inherent upper bound on prediction accuracy for older age groups that cannot be overcome through architectural improvements alone, since the biological signal itself becomes less reliable. Importantly, the age groups spanning 20 to 65 years, which constitute 55.2\% of test samples, achieve MAE values between 3.5 and 4.2 years for our method, remaining close to the overall test MAE of 3.149 and confirming that the model generalises well across the most clinically relevant age range.

\paragraph{Alignment with chronological age.}
The scatter plots in Figure~\ref{fig:scatter} illustrate a
qualitative difference in model behaviour that is not fully
reflected by the MAE values alone. The sequence-agnostic
baseline demonstrates a notable regression-to-the-mean
effect, as evidenced by a fit line slope of 0.905,
indicating systematic underestimation of age in older
individuals and overestimation in younger ones. This is a
characteristic failure mode arising when node
representations lack adequate discriminative power to
differentiate the full dynamic range of methylation states.
Without sequence context, the integrated framework cannot
determine which CpG sites carry age-driven methylation
signals versus those whose variation is attributable to
individual differences unrelated to aging, resulting in
conservative predictions that regress toward the training
mean. CNN-based sequence encoding partially mitigates this,
raising the slope to 0.946, though improvement is limited
by the CNN's inability to reliably extract informative
sequence features at this dataset scale. Our statistical
feature approach achieves the highest slope of 0.954,
approaching the ideal value of 1.0, and produces the most
compact point cloud around the diagonal. The $R^2$ values
follow the same ordering (0.9491, 0.9615, 0.9619),
confirming that jointly integrating sequence context and
graph structure enhances the model's capacity to capture
the full dynamic range of epigenetic age.

\subsection{Sequence Feature Interpretability via GNNExplainer}
\label{sec:interpretability}

\begin{figure}[h]
\centering
\includegraphics[width=\linewidth]{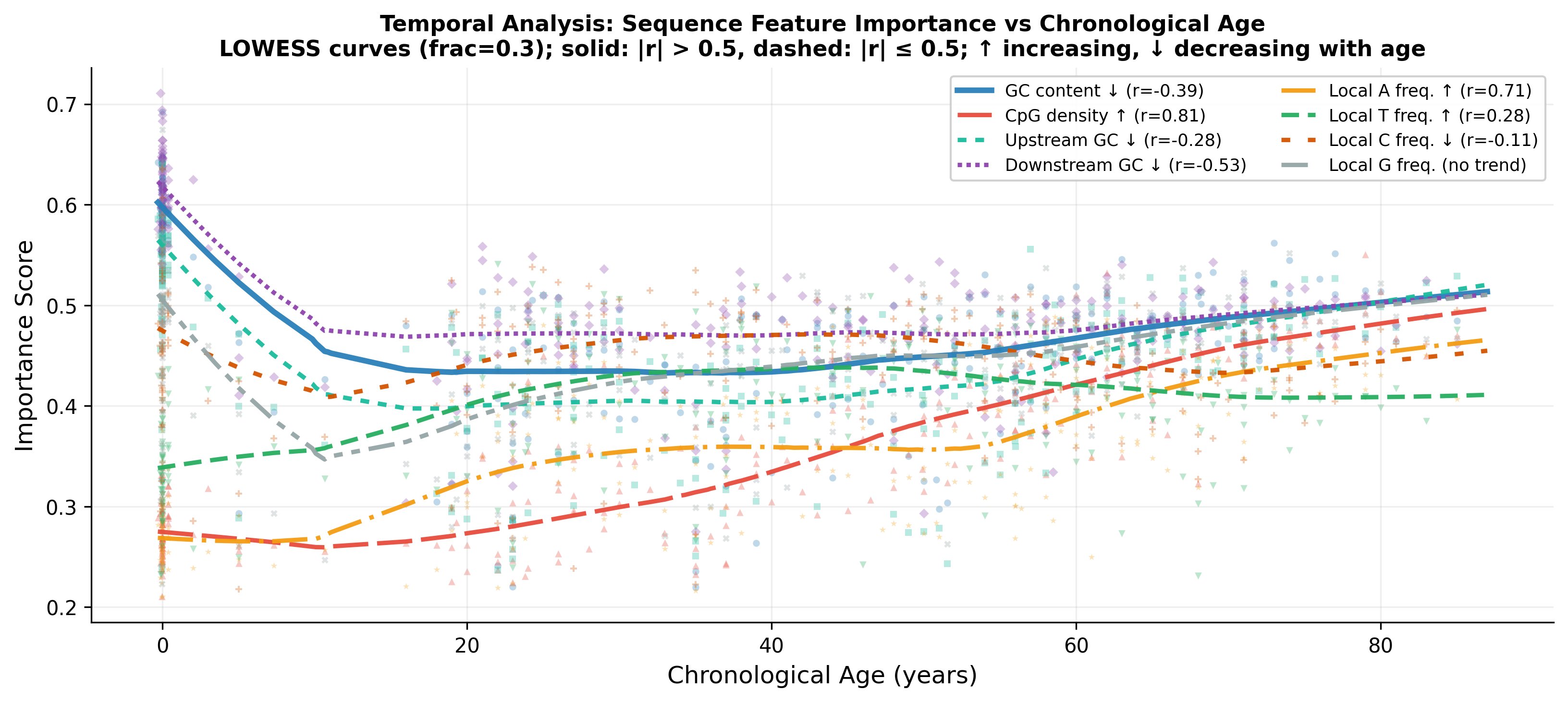}
\caption{Temporal analysis of sequence feature importance
         scores.}
\label{fig:seq_temporal}
\end{figure}

\paragraph{Sequence feature importance across the age spectrum.}
Figure~\ref{fig:seq_temporal} presents the temporal
analysis of sequence feature importance scores across
chronological age, using LOWESS regression to
capture non-linear trends without imposing a parametric
form. Pearson correlation coefficients are computed to
characterise the overall direction and strength of each
trend; features with $p > 0.01$ are shown without a
smoothed curve. Linear regression results are provided
in Appendix~\ref{app:linear_temporal} for comparison.

Two features exhibit increasing importance across age groups, from younger to older individuals: CpG density ($r = 0.81$, $p < 0.001$) and local
adenine frequency ($r = 0.71$, $p < 0.001$). The LOWESS
curves reveal that this increase is non-linear: both
features show a plateau or slight decline from childhood through adolescence (0--20 years), followed by a sustained
increase through adulthood into old age. This pattern
suggests that sequence-determined regulatory context
becomes progressively more relevant as stochastic
epigenetic drift accumulates, consistent with the known
biology of aging: CpG density is a primary determinant of
CpG island regulatory context, and age-related
hypermethylation preferentially targets CpG-dense promoter
regions~\cite{issa2000}.

Downstream GC content ($r = -0.53$, $p < 0.001$) exhibits
a distinct non-linear decline: the LOWESS curve shows a
sharp decrease in importance during the first decade of
life, followed by stabilisation through adulthood. This
suggests that flanking sequence composition is most
discriminative in early development, where methylation
patterns are tightly coupled to sequence context, but
becomes less informative as epigenetic drift increasingly
dominates in older subjects. Notably, all features exhibit
elevated importance scores near birth, reflecting the
distinctly structured methylation landscape of neonatal
samples relative to older age groups. Local guanine
frequency shows no significant trend ($p = 0.88$),
contributing uniformly across the lifespan.

Site-level importance trajectories for individual CpG sites are shown in 
Appendix~\ref{app:cpg_temporal}. Full details of the GNNExplainer~\cite{ying2019} aggregation procedure are provided in Appendix~\ref{app:explainer}.

\section{Conclusion and Limitations}
\paragraph{Conclusion.}
We proposed a unified sequence--graph integration framework
for epigenetic age prediction that jointly models
co-methylation graph structure and site-specific DNA
sequence context via a lightweight gated modulation
mechanism. Our framework achieves a test MAE of 3.149 years,
surpassing all evaluated baselines including classical
epigenetic clocks, deep learning age predictors, and
graph-based methods, with a 12.8\% improvement over the
strongest graph-based baseline. Biologically informed statistical
features outperform CNN-based sequence encoding, and
post-hoc interpretability analysis reveals age-dependent
importance shifts in CpG density and local adenine
frequency consistent with known aging biology.

\paragraph{Limitations.}
All experiments are conducted on blood methylation samples
from a single array platform 
(GPL8490\footnote{\url{https://www.ncbi.nlm.nih.gov/geo/query/acc.cgi?acc=GPL8490}}), and generalisation
to other tissue types or higher-density arrays such as the
450K or EPIC platforms remains to be validated. Training
is performed on a single cross-validation fold rather than
averaging across all folds, which limits the statistical
robustness of the reported performance estimates. The
eight-dimensional sequence statistical features are
derived from the reference genome and are therefore
invariant across individuals, meaning the model cannot
capture individual-specific sequence variation such as
single nucleotide polymorphisms that may influence local
methylation susceptibility. Finally, the GNNExplainer~\cite{ying2019}
interpretability analysis assigns importance scores in a
post-hoc manner and reflects correlations between sequence
features and model predictions rather than causal
relationships. The identified age-dependent importance
shifts in CpG density and local adenine frequency should
therefore be treated as hypothesis-generating observations
that warrant further experimental validation.

\FloatBarrier

\bibliographystyle{unsrtnat}
\bibliography{references}

@article{horvath2013,
  author  = {Horvath, Steve},
  title   = {{DNA} methylation age of human tissues and cell types},
  journal = {Genome Biology},
  year    = {2013},
  volume  = {14},
  number  = {10},
  pages   = {R115},
  doi     = {10.1186/gb-2013-14-10-r115}
}

@article{hannum2013,
  author  = {Hannum, Gregory and Guinney, Justin and Zhao, Ling and
             Zhang, Li and Hughes, Guy and Sadda, SriniVas and
             Klotzle, Brandy and Bibikova, Marina and Fan, Jian-Bing and
             Gao, Yuan and Deconde, Rob and Chen, Menzies and
             Rajapakse, Indika and Friend, Stephen and Ideker, Trey and
             Zhang, Kang},
  title   = {Genome-wide methylation profiles reveal quantitative views
             of human aging rates},
  journal = {Molecular Cell},
  year    = {2013},
  volume  = {49},
  number  = {2},
  pages   = {359--367},
  doi     = {10.1016/j.molcel.2012.10.016}
}

@article{galkin2021,
  author  = {Galkin, Fedor and Mamoshina, Polina and Kochetov, Kirill and
             Sidorenko, Denis and Zhavoronkov, Alex},
  title   = {{DeepMAge}: A methylation aging clock developed with deep learning},
  journal = {Aging and Disease},
  year    = {2021},
  volume  = {12},
  number  = {5},
  pages   = {1252--1262},
  doi     = {10.14336/AD.2020.1202}
}

@article{delimacamillo2022,
  author  = {de Lima Camillo, Lucas Paulo and Lapierre, Louis R. and Singh, Ritambhara},
  title   = {A pan-tissue {DNA}-methylation epigenetic clock based on deep learning},
  journal = {npj Aging},
  year    = {2022},
  volume  = {8},
  number  = {1},
  pages   = {4},
  doi     = {10.1038/s41514-022-00085-y}
}

@article{wu2019,
  author  = {Wu, Xiangyu and Wei, Zhen and Chen, Kunqi and Zhang, Qing and
             Su, Jionglong and Liu, Hui and Zhang, Lin and Meng, Jia},
  title   = {m6{A}comet: large-scale functional prediction of individual
             m6{A} {RNA} methylation sites from an {RNA} co-methylation network},
  journal = {BMC Bioinformatics},
  year    = {2019},
  volume  = {20},
  number  = {1},
  pages   = {223},
  doi     = {10.1186/s12859-019-2840-3}
}

@article{santoni2021,
  author  = {Santoni, Daniele},
  title   = {The impact of flanking sequence features on {DNA} {CpG} methylation},
  journal = {Computational Biology and Chemistry},
  year    = {2021},
  volume  = {92},
  pages   = {107480},
  doi     = {10.1016/j.compbiolchem.2021.107480}
}

@article{affinito2020,
  author  = {Affinito, Ornella and Palumbo, Domenico and Fierro, Annalisa and
             Cuomo, Mariella and De Riso, Giulia and Monticelli, Antonella and
             Miele, Gennaro and Chiariotti, Lorenzo and Cocozza, Sergio},
  title   = {Nucleotide distance influences co-methylation between nearby {CpG} sites},
  journal = {Genomics},
  year    = {2020},
  volume  = {112},
  number  = {1},
  pages   = {144--150},
  doi     = {10.1016/j.ygeno.2019.05.007}
}

@article{shi2024,
  author  = {Shi, Lijuan and Hai, Boquan and Kuang, Zhejun and Wang, Han and Zhao, Jian},
  title   = {{ResnetAge}: A Resnet-based {DNA} methylation age prediction method},
  journal = {Bioengineering},
  year    = {2024},
  volume  = {11},
  number  = {1},
  pages   = {34},
  doi     = {10.3390/bioengineering11010034}
}

@article{prosz2024,
  author  = {Prosz, Aurel and Pipek, Orsolya and B{\"o}rcs{\"o}k, Judit and
             Palla, Gergely and Szallasi, Zoltan and Spisak, Sandor and Csabai, Istvan},
  title   = {Biologically informed deep learning for explainable epigenetic clocks},
  journal = {Scientific Reports},
  year    = {2024},
  volume  = {14},
  pages   = {1306},
  doi     = {10.1038/s41598-023-50495-5}
}

@article{ahmed2025,
  author  = {Ahmed, Saleh Sakib and Shabab, Nahian and
             Samee, Md Abul Hassan and Rahman, M. Sohel},
  title   = {{GraphAge}: Unleashing the power of graph neural network
             to decode epigenetic aging},
  journal = {PNAS Nexus},
  year    = {2025},
  volume  = {4},
  number  = {6},
  pages   = {pgaf177},
  doi     = {10.1093/pnasnexus/pgaf177}
}

@inproceedings{corso2020,
  author    = {Corso, Gabriele and Cavalleri, Luca and Beaini, Dominique and
               Li{\`o}, Pietro and Veli{\v{c}}kovi{\'{c}}, Petar},
  title     = {Principal neighbourhood aggregation for graph nets},
  booktitle = {Advances in Neural Information Processing Systems},
  year      = {2020},
  volume    = {33},
  pages     = {13260--13271}
}

@article{angermueller2017,
  author  = {Angermueller, Christof and Lee, Heather J. and Reik, Wolf and Stegle, Oliver},
  title   = {{DeepCpG}: accurate prediction of single-cell {DNA} methylation
             states using deep learning},
  journal = {Genome Biology},
  year    = {2017},
  volume  = {18},
  pages   = {67},
  doi     = {10.1186/s13059-017-1189-z}
}

@article{lu2023,
  author  = {Lu, Yuancheng Ryan and Tian, Xiao and Sinclair, David A.},
  title   = {The information theory of aging},
  journal = {Nature Aging},
  year    = {2023},
  volume  = {3},
  number  = {12},
  pages   = {1486--1499},
  doi     = {10.1038/s43587-023-00527-6}
}

@inproceedings{ying2019,
  author    = {Ying, Zhitao and Bourgeois, Dylan and You, Jiaxuan and
               Zitnik, Marinka and Leskovec, Jure},
  title     = {{GNNExplainer}: Generating explanations for graph neural networks},
  booktitle = {Advances in Neural Information Processing Systems},
  year      = {2019},
  volume    = {32},
  pages     = {9240--9251}
}

@article{issa2000,
  author  = {Issa, Jean-Pierre},
  title   = {{CpG}-island methylation in aging and cancer},
  journal = {Current Topics in Microbiology and Immunology},
  year    = {2000},
  volume  = {249},
  pages   = {101--118},
  doi     = {10.1007/978-3-642-59696-4_7}
}

@article{lopezotin2013,
  author  = {L{\'o}pez-Ot{\'i}n, Carlos and Blasco, Maria A. and 
             Partridge, Linda and Serrano, Manuel and Kroemer, Guido},
  title   = {The hallmarks of aging},
  journal = {Cell},
  year    = {2013},
  volume  = {153},
  number  = {6},
  pages   = {1194--1217},
  doi     = {10.1016/j.cell.2013.05.039}
}

@article{horvathraj2018,
  author  = {Horvath, Steve and Raj, Kenneth},
  title   = {{DNA} methylation-based biomarkers and the epigenetic 
             clock theory of ageing},
  journal = {Nature Reviews Genetics},
  year    = {2018},
  volume  = {19},
  number  = {6},
  pages   = {371--384},
  doi     = {10.1038/s41576-018-0004-3}
}

@article{jaenischbird2003,
  author  = {Jaenisch, Rudolf and Bird, Adrian},
  title   = {Epigenetic regulation of gene expression: how the genome 
             integrates intrinsic and environmental signals},
  journal = {Nature Genetics},
  year    = {2003},
  volume  = {33},
  number  = {Suppl},
  pages   = {245--254},
  doi     = {10.1038/ng1089}
}

@article{levine2018,
  author  = {Levine, Morgan E. and Lu, Ake T. and Quach, Austin and
             Chen, Brian H. and Assimes, Themistocles L. and 
             Bandinelli, Stefania and Hou, Lifang and 
             Baccarelli, Andrea A. and Stewart, James D. and 
             Li, Yun and Whitsel, Eric A. and Wilson, James G. and
             Reiner, Alex P. and Aviv, Abraham and Lohman, Kurt and
             Liu, Yongmei and Ferrucci, Luigi and Horvath, Steve},
  title   = {An epigenetic biomarker of aging for lifespan 
             and healthspan},
  journal = {Aging},
  year    = {2018},
  volume  = {10},
  number  = {4},
  pages   = {573--591},
  doi     = {10.18632/aging.101414}
}

@article{schubeler2015,
  author  = {Sch{\"u}beler, Dirk},
  title   = {Function and information content of {DNA} methylation},
  journal = {Nature},
  year    = {2015},
  volume  = {517},
  number  = {7534},
  pages   = {321--326},
  doi     = {10.1038/nature14192}
}

@article{gao2020,
  author  = {Gao, Linfeng and Emperle, Max and Guo, Yiran and
             Grimm, Sara A. and Ren, Wendan and Adam, Sabrina and
             Uryu, Hidetaka and Zhang, Zhi-Min and Chen, Dongliang and
             Yin, Jiekai and Dukatz, Michael and Anteneh, Hiwot and
             Jurkowska, Renata Z. and Lu, Jiuwei and Wang, Yinsheng and
             Bashtrykov, Pavel and Wade, Paul A. and Wang, Gang Greg and
             Jeltsch, Albert and Song, Jikui},
  title   = {Comprehensive structure-function characterization of
             {DNMT3B} and {DNMT3A} reveals distinctive de novo
             {DNA} methylation mechanisms},
  journal = {Nature Communications},
  year    = {2020},
  volume  = {11},
  number  = {1},
  pages   = {3355},
  doi     = {10.1038/s41467-020-17109-4}
}

@article{deaton2011,
  author  = {Deaton, Aim{\'e}e M. and Bird, Adrian},
  title   = {{CpG} islands and the regulation of transcription},
  journal = {Genes \& Development},
  year    = {2011},
  volume  = {25},
  number  = {10},
  pages   = {1010--1022},
  doi     = {10.1101/gad.2037511}
}

@article{issa2014,
  author  = {Issa, Jean-Pierre},
  title   = {Aging and epigenetic drift: a vicious cycle},
  journal = {Journal of Clinical Investigation},
  year    = {2014},
  volume  = {124},
  number  = {1},
  pages   = {24--29},
  doi     = {10.1172/JCI69735}
}

@article{jones2015,
  author  = {Jones, Meaghan J. and Goodman, Sarah J. and Kobor, Michael S.},
  title   = {{DNA} methylation and healthy human aging},
  journal = {Aging Cell},
  year    = {2015},
  volume  = {14},
  number  = {6},
  pages   = {924--932},
  doi     = {10.1111/acel.12349}
}

@inproceedings{kingma2015,
  author    = {Kingma, Diederik P. and Ba, Jimmy},
  title     = {Adam: A method for stochastic optimization},
  booktitle = {International Conference on Learning Representations},
  year      = {2015}
}

@article{srivastava2014,
  author  = {Srivastava, Nitish and Hinton, Geoffrey and 
             Krizhevsky, Alex and Sutskever, Ilya and 
             Salakhutdinov, Ruslan},
  title   = {Dropout: A simple way to prevent neural networks 
             from overfitting},
  journal = {Journal of Machine Learning Research},
  year    = {2014},
  volume  = {15},
  pages   = {1929--1958}
}

@inproceedings{klambauer2017,
  author    = {Klambauer, G{\"u}nter and Unterthiner, Thomas and 
               Mayr, Andreas and Hochreiter, Sepp},
  title     = {Self-normalizing neural networks},
  booktitle = {Proceedings of the 31st International Conference on 
               Neural Information Processing Systems},
  series    = {NIPS'17},
  year      = {2017},
  pages     = {972--981},
  publisher = {Curran Associates Inc.},
  address   = {Red Hook, NY, USA}
}

@article{lopezotin2023,
  author  = {L{\'o}pez-Ot{\'i}n, Carlos and Blasco, Maria A. and 
             Partridge, Linda and Serrano, Manuel and Kroemer, Guido},
  title   = {Hallmarks of aging: An expanding universe},
  journal = {Cell},
  year    = {2023},
  volume  = {186},
  number  = {2},
  pages   = {243--278},
  doi     = {10.1016/j.cell.2022.11.001}
}

@article{ektefaie2023,
  author  = {Ektefaie, Yasha and Dasoulas, George and Noori, Ayush and 
             Farhat, Maha and Zitnik, Marinka},
  title   = {Multimodal learning with graphs},
  journal = {Nature Machine Intelligence},
  year    = {2023},
  volume  = {5},
  number  = {4},
  pages   = {340--350},
  doi     = {10.1038/s42256-023-00624-6}
}

@inproceedings{nguyen2023,
  author    = {Nguyen, Eric and Poli, Michael and Faizi, Marjan and
               Thomas, Armin W. and Sykes, Callum Birch and
               Wornow, Michael and Patel, Aman and Rabideau, Clayton and
               Massaroli, Stefano and Bengio, Yoshua and Ermon, Stefano and
               Baccus, Stephen A. and R{\'e}, Christopher},
  title     = {{HyenaDNA}: long-range genomic sequence modeling at
               single nucleotide resolution},
  booktitle = {Proceedings of the 37th International Conference on
               Neural Information Processing Systems},
  series    = {NIPS '23},
  year      = {2023},
  publisher = {Curran Associates Inc.},
  address   = {Red Hook, NY, USA}
}

@inproceedings{zhou2024,
  author    = {Zhou, Zhihan and Ji, Yanrong and Li, Weijian and
               Dutta, Pratik and Davuluri, Ramana V. and Liu, Han},
  title     = {{DNABERT-2}: Efficient foundation model and benchmark
               for multi-species genomes},
  booktitle = {International Conference on Learning Representations},
  year      = {2024}
}

@inproceedings{schiff2024,
  author    = {Schiff, Yair and Kao, Chia-Hsiang and Gokaslan, Aaron and
               Dao, Tri and Gu, Albert and Kuleshov, Volodymyr},
  title     = {Caduceus: bi-directional equivariant long-range {DNA}
               sequence modeling},
  booktitle = {Proceedings of the 41st International Conference on
               Machine Learning},
  series    = {ICML'24},
  year      = {2024},
  publisher = {JMLR.org}
}

@inproceedings{digiovanni2023,
  author    = {Di Giovanni, Francesco and Giusti, Lorenzo and 
               Barbero, Federico and Luise, Giulia and 
               Li{\`o}, Pietro and Bronstein, Michael},
  title     = {On over-squashing in message passing neural networks:
               the impact of width, depth, and topology},
  booktitle = {Proceedings of the 40th International Conference on
               Machine Learning},
  series    = {ICML'23},
  year      = {2023},
  publisher = {JMLR.org}
}

@inproceedings{hu2024,
  author    = {Hu, Bozhen and Tan, Cheng and Xia, Jun and Liu, Yue and
               Wu, Lirong and Zheng, Jiangbin and Xu, Yongjie and
               Huang, Yufei and Li, Stan Z.},
  title     = {Learning complete protein representation by dynamically
               coupling of sequence and structure},
  booktitle = {Advances in Neural Information Processing Systems},
  volume    = {37},
  pages     = {137673--137697},
  publisher = {Curran Associates, Inc.},
  year      = {2024}
}

@inproceedings{marin2024,
  author    = {Marin, Frederikke and Teufel, Felix and Horlacher, Marc and
               Madsen, Dennis and Pultz, Dennis and Winther, Ole and
               Boomsma, Wouter},
  title     = {{BEND}: Benchmarking {DNA} language models on
               biologically meaningful tasks},
  booktitle = {International Conference on Learning Representations},
  year      = {2024},
  pages     = {15246--15281}
}

@misc{ncbi2024,
  author       = {{National Center for Biotechnology Information}},
  title        = {{NCBI}},
  year         = {2024},
  url          = {https://www.ncbi.nlm.nih.gov/}
}

@misc{ebi2024,
  author       = {{EMBL-EBI}},
  title        = {{EMBL's European Bioinformatics Institute}},
  year         = {2024},
  url          = {https://www.ebi.ac.uk/}
}


\appendix

\section{Dataset Details and Preprocessing}
\label{app:dataset}

\paragraph{Sample filtering.}
Each dataset is filtered to retain only blood tissue samples.
Samples with any missing methylation values are excluded.
The resulting samples are restricted to the 20,318 CpG sites
present in the AltumAge multi-platform CpG
list~\cite{delimacamillo2022}. All train/test splits are
performed prior to any graph construction to prevent data
leakage.

\paragraph{CpG site annotation preprocessing.}
Positional and biological attributes for each CpG site are
loaded from the GPL8490 Illumina HumanMethylation27 manifest.
Sites with missing chromosome assignments are removed. CpG
island start and end positions are parsed from the island
location field, with missing values defaulting to zero. CpG
island length is computed as the difference between end and
start positions. The following continuous attributes are
normalised per chromosome using Min-Max scaling: genomic
position, TSS coordinate, CpG island start and end
positions, CpG island length, and distance to TSS. Missing
distance-to-TSS values are filled with the maximum
normalised value of 1 after scaling. The following
categorical attributes are one-hot encoded: gene strand
direction, next adjacent nucleotide, and source strand
orientation.

\paragraph{Node feature construction.}
Each CpG site node carries a ten-dimensional feature vector.
The first dimension is the sample-varying methylation beta
value. The remaining nine dimensions are fixed positional
attributes derived from the preprocessed manifest: CpG
island membership, CpG island length, distance to TSS,
next-base encoding spanning three dimensions, normalised
CpG island start position, normalised CpG island end
position, and normalised genomic coordinate.

\paragraph{Graph construction.}
The co-methylation graph is constructed separately for each
cross-validation fold using only the fold's training
samples. This ensures no information from the validation or test
sets influences the graph topology. Pearson correlation
coefficients are computed across all pairs of 20,318 CpG
sites using training-fold methylation beta values. An edge
between sites $i$ and $j$ is included under any of the
following conditions, with self-loops excluded. First, the
absolute correlation $|r_{ij}|$ exceeds 0.70, regardless
of chromosome. Second, sites $i$ and $j$ are on the same
chromosome and $|r_{ij}|$ exceeds 0.68. Third, sites $i$
and $j$ are on the same chromosome, their genomic distance
is less than $10^{5}$\,bp, and $|r_{ij}|$ exceeds 0.66.
Each edge carries a three-dimensional attribute vector
encoding the Pearson correlation coefficient, a
same-chromosome binary indicator, and a same-gene binary
indicator.

\paragraph{DNA sequence feature extraction.}
The 122\,bp surrounding genomic sequence for each CpG site
is retrieved from the platform manifest. The CpG
dinucleotide marker is standardised and all characters are
uppercased prior to feature extraction. Unknown nucleotides
are assigned equal probability across all four bases. All
sequence features are invariant across samples and are
computed once prior to training. The eight-dimensional
statistical feature vector is described in
Table~\ref{tab:seq_features}.

\begin{figure}[H]
\centering
\includegraphics[width=0.8\textwidth]{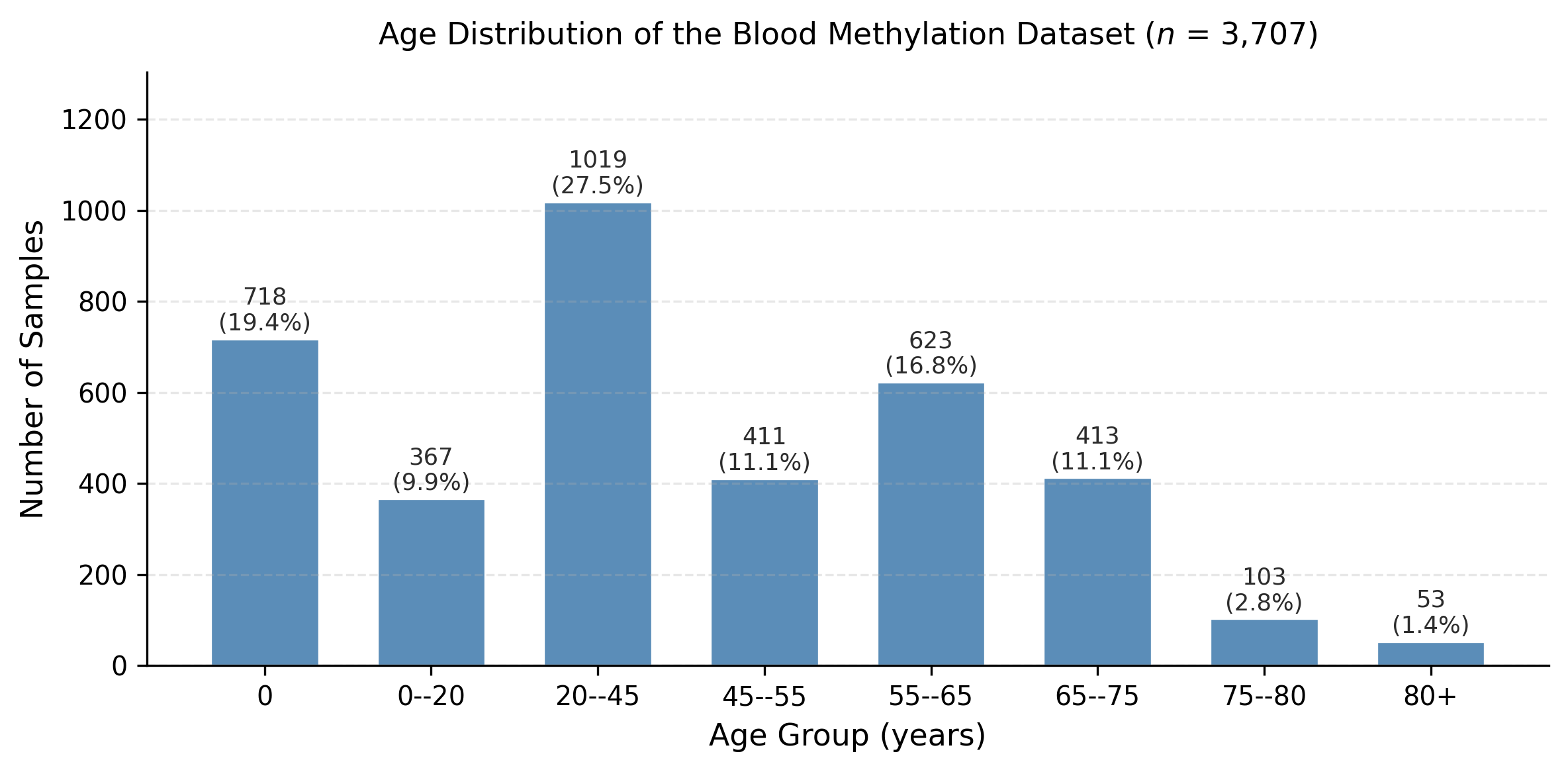}
\caption{Age distribution of the blood methylation dataset ($n=3{,}707$).}
\label{fig:age_dist}
\end{figure}

\section{Hardware and Training Details}
\label{app:hardware}

All experiments are conducted on a single NVIDIA H100 GPU.
Total training time for fold~2 is approximately 9 hours.
A batch size of 1 is used throughout training. A fixed
random seed of 0 is applied to Python, NumPy, and PyTorch
to ensure reproducibility.

\section{Linear Regression Temporal Analysis}
\label{app:linear_temporal}

\begin{figure}[H]
\centering
\includegraphics[width=0.8\textwidth]{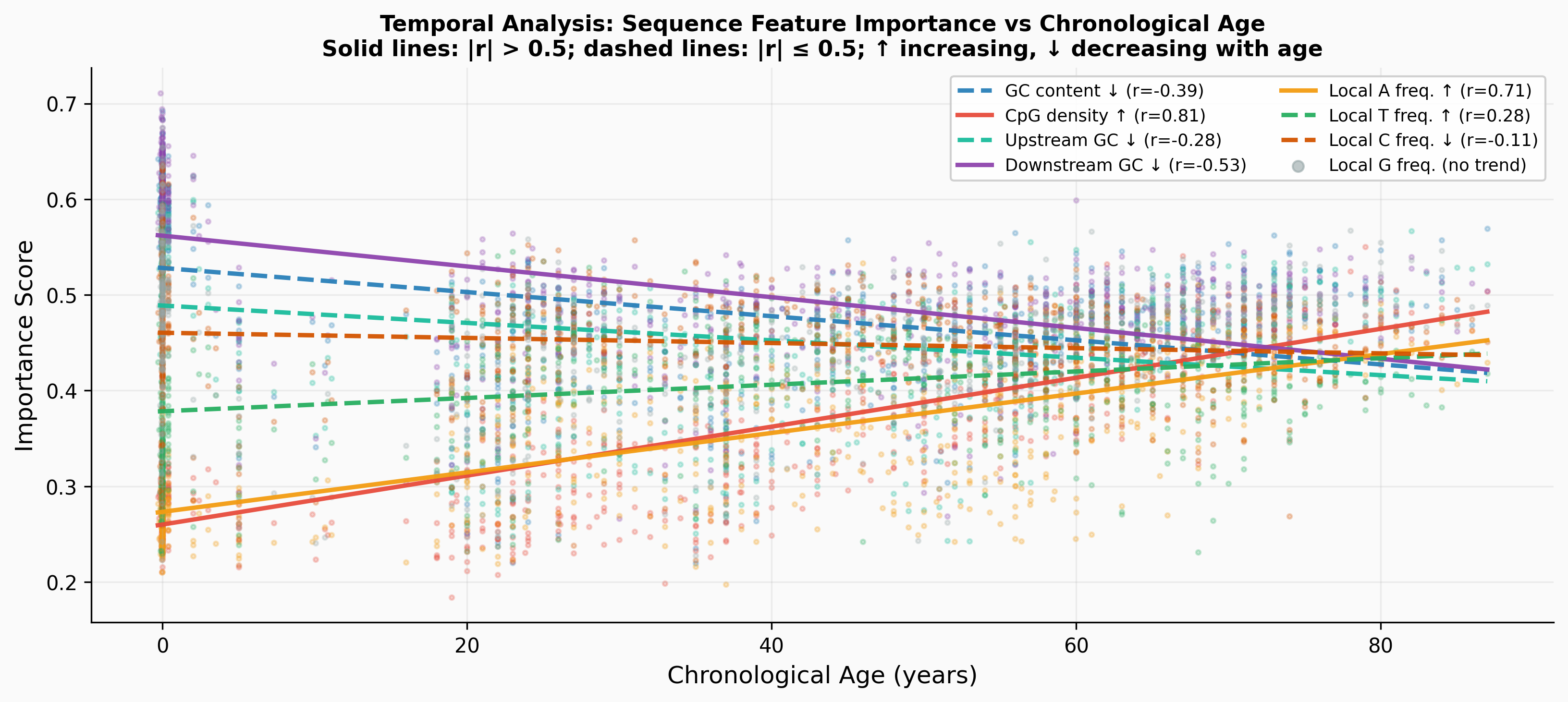}
\caption{Linear regression fits of sequence feature importance scores 
         across chronological age.}
\label{fig:seq_linear}
\end{figure}

Figure~\ref{fig:seq_linear} presents linear regression fits
of sequence feature importance scores across chronological
age, complementing the LOWESS analysis in
Figure~\ref{fig:seq_temporal}. The linear fits confirm the
directional trends identified by LOWESS. CpG density
exhibits a statistically significant positive slope with
$r = 0.81$. Local adenine frequency also shows a
significant positive slope with $r = 0.71$. Downstream GC
content shows a significant negative slope with $r = -0.53$.
GC content and upstream GC content show weak negative trends
with $r = -0.39$ and $r = -0.28$ respectively. Local
thymine frequency shows a weak positive trend with
$r = 0.28$. Local guanine and cytosine frequencies show no
significant linear trend.

\section{CpG Site-Level Importance Trajectories}
\label{app:cpg_temporal}

\begin{figure}[H]
\centering
\includegraphics[width=\textwidth]{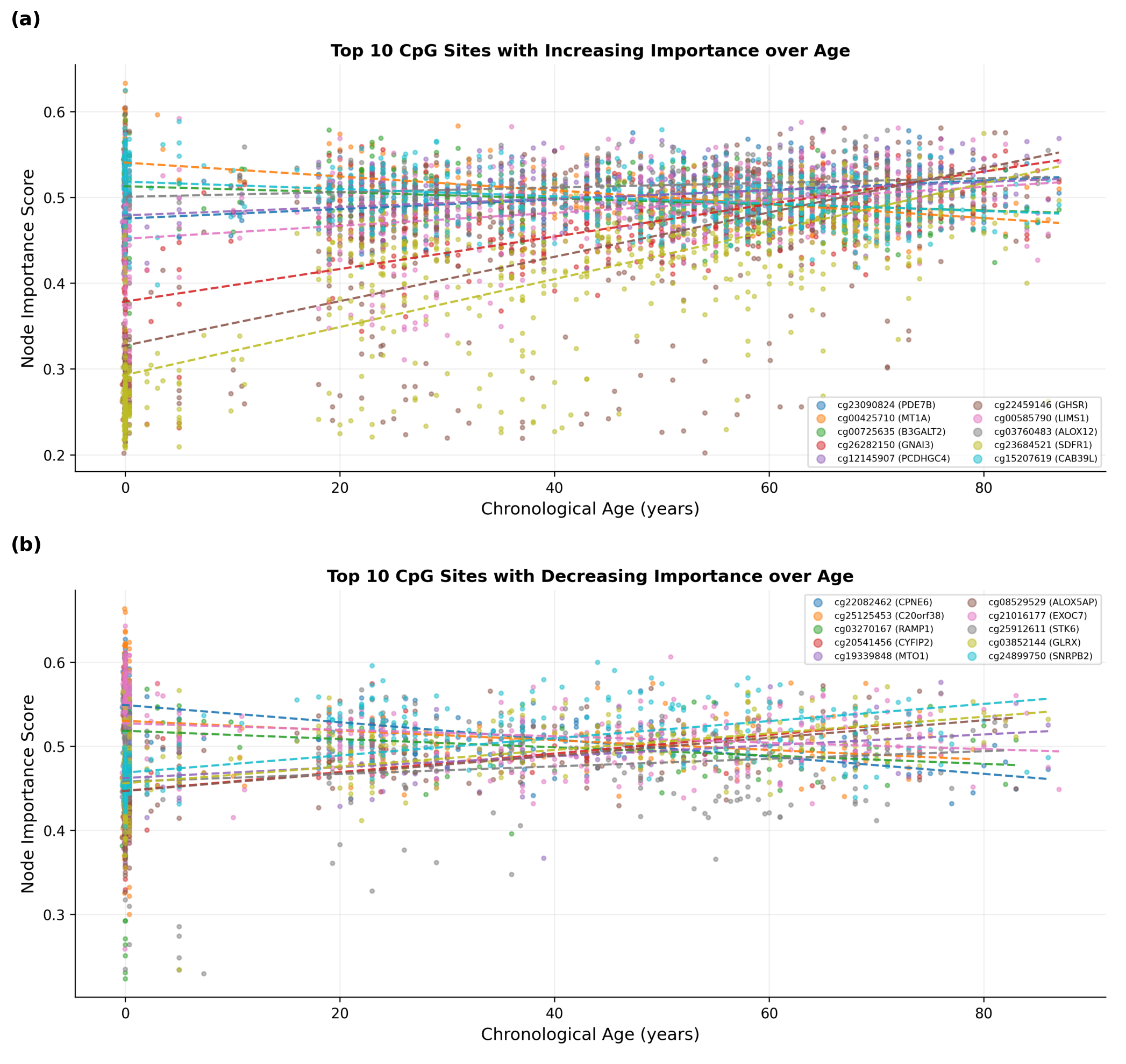}
\caption{Top 10 CpG sites with increasing (a) and decreasing (b)
         node importance scores across chronological age,
         fitted with linear regression.}
\label{fig:cpg_temporal}
\end{figure}

Figure~\ref{fig:cpg_temporal} presents node-level importance
trajectories for the top 10 CpG sites with increasing and
decreasing importance across chronological age. Sites with increasing
importance are associated with genes involved in neuronal
signalling and metabolic regulation. This suggests that
sequence-regulatory context at these loci becomes
progressively more discriminative for age prediction with
advancing age. Sites with decreasing importance are
associated with genes implicated in cytoskeletal
organisation and mitochondrial function. This is consistent
with their methylation patterns being most informative
during early development and less so in older subjects.

\section{GNNExplainer Aggregation Details}
\label{app:explainer}

Post-hoc interpretability analysis is conducted using
GNNExplainer~\cite{ying2019} applied to all 756 test
samples independently. Two complementary analyses are
performed.

\paragraph{Sequence feature importance.}
GNNExplainer is applied with a shared attribute mask
setting, which directly produces a single eight-dimensional
importance vector that is common across all 20,318 CpG
site nodes within each sample. This formulation does not
require aggregation across nodes. The importance vector
reflects the contribution of each sequence statistical
feature to the model prediction at the graph level.
Per-sample importance vectors are averaged across all 756
test samples to obtain the mean importance score for each
feature. For the temporal analysis in
Figure~\ref{fig:seq_temporal}, the per-sample importance
vectors are regressed against the corresponding
chronological age using Pearson correlation to
characterise age-dependent importance trends.

\paragraph{CpG site-level importance.}
GNNExplainer is applied with a per-node scalar mask
setting, producing an independent scalar importance score
for each of the 20,318 CpG site nodes per sample. This
yields a matrix of shape $756 \times 20{,}318$ containing
per-sample per-node importance scores. For each CpG site,
a linear regression is fitted between its importance
scores across all 756 samples and the corresponding
chronological ages. The resulting slope is used to rank
sites by their age-dependent importance trend.



\end{document}